# A Deep Learning approach for Hindi Named Entity Recognition

**Bansi Shah**
Indian Institute of Technology
Kharagpur

**Sunil Kumar Kopparapu**
TCS Research and Innovation - Mumbai
Yantra Park Thane (West), Maharashtra

## Abstract

Named Entity Recognition is one of the most important text processing requirement in many NLP tasks. In this paper we use a deep architecture to accomplish the task of recognizing named entities in a given Hindi text sentence. Bidirectional Long Short Term Memory (BiLSTM) based techniques have been used for NER task in literature. In this paper, we first tune BiLSTM low-resource scenario to work for Hindi NER and propose two enhancements namely (a) de-noising auto-encoder (DAE) LSTM and (b) conditioning LSTM which show improvement in NER task compared to the BiLSTM approach. We use pre-trained word embedding to represent the words in the corpus, and the NER tags of the words are as defined by the used annotated corpora. Experiments have been performed to analyze the performance of different word embeddings and batch sizes which is essential for training deep models.

## 1 Introduction

The task of Named Entity Recognition (NER) was coined in 1995 in Message Understanding Conference-6 (MUC-6). It is defined to be consisting of three subtasks, namely, entity names, temporal expressions and number expressions. The expressions to be annotated are "unique identifiers" of (a) entities like names of organizations, names of persons or names of locations, (b) temporal like dates and times, and (c) quantities like monetary values, percentages (MUC-6, 1995). As can be guessed, NER is one of the key tasks in the field of information extraction and Natural Language Processing (NLP). English language can boast of a rich NER literature, however, the same can not be said to be true for Hindi language. While there have been sporadic attempts, there is much to be explored in the Hindi language domain, especially considering that use of deep learning models have made their way into several language processing problems. Lack of ready tools, rich morphology nature of Hindi language and more importantly scarcity of annotated corpus makes (a) reusing existing deep learning architectures used for English language challenging and (b) allows exploring novel and interesting approaches in general and specifically for NER task.

In this paper, based on the success of using machine learning architectures for NER task, for resource rich languages like English, we follow a simple yet effective approach of refining previously proven successful deep neural network models for Hindi language. The idea is to use sparse deep neural network architecture which allows to learn the model parameters in low-resource scenario. The architecture geared towards low-resource data has also the advantage that it allows not only using less resources in terms of computing time and power but also shows an improvement over the existing models for the Hindi NER task. Specifically, the main contribution of this paper is the use of two basic learning architectures in an hierchical stack; the first model (BiLSTM) in the stack helps estimate an initial NER output which is then fed to the second model in the stack to obtain an improved NER over the NER output given by the first model in the stack. Using this kind of hierchical architecture, we show experimentally that there is an improvement in Hindi NER performance over the base BiLSTM model by appending a small amount of network model parameters to the base BiLSTM model architecture. We believe that these kind of modifications or integration of different network models help improve Hindi NER performance especially in low-resource conditions.

The rest of the paper is organized as follow. In Section 2 we survey existing work in Hindi NER including the use of deep learning architecture and introduce word embeddings which is central to the representation of words in most NLP tasks. In Section 3 we introduce the dataset and describe the proposed hirerchical learning approaches for Hindi NER in Section 4. Experimental results and analysis is discussed in Section 5 and we conclude in Section 6.

## 2 Related Work

While NER has a rich literature, it was not until 2008, a lot of work for NER on Indian languages saw prominence. Initially, NER task was based on language specific designing rules and gazetteer lists. They designed language specific rules, made gazetteer lists to add knowledge to the data and performed NER on Indian Languages. Many approaches were proposed, for example, (Saha et al., 2008) used class specific language rules, gazetteer as features to their Maximum Entropy Model. (Ekbal and Bandyopadhyay, 2008) proposed pair wise multi-class decision method and second degree polynomial kernel function to perform classification on the text data using SVM. Another approach using CRFs (Conditional Random Field) (Gali et al., 2008) used rules similar to previous findings for designing the CRF features in addition to gazetteer lists. In 2013, (Das and Garain, 2014) also used gazetteer lists and linguistic features for classification of the text using a CRF model.

With the advent of Deep Learning, in 2016 (Athavale et al., 2016) proposed a technique to identify named entities in a given piece of text without any language specific rules. They used a bidirectional Long Short Term Memory (BiLSTM) model that classified the words in the sentences into one of the required classes. Their model significantly outperformed previous approaches involving rule based systems or hand-crafted features. In 2018, (Xie et al., 2018) propose a method that translated the low-resource language to a resource rich language and then using the tools available for the rich language to perform a NER on the translated text. This approach makes use of the fact that there is a good language translation from the low-resource to rich resource language, which often is not the case. A model that combined deep learning architecture with knowledge-based feature extractors was explored in (Dadas, 2018). They use a vectorized representation of the word, which is constructed by concatenating (a) the output of a pre-trained word embedding, (b) a train-able character level encoder and (c) a set of one-hot vectors from feature extraction module. Then a hidden word representation is computed by a number of BiLSTM layers. Finally, this representation is sent to a CRF output layer, which is responsible for predicting a sequence of labels for all the words in the sentence.

We use the (Athavale et al., 2016) BiLSTM architecture as our initial model, which we also use as our base model to compare our proposed NER models. We propose a method to use auto-encoders and conditioning LSTM's on top of this model to improve the performance of the NER task for Hindi language. Word embeddings are representation of the words in any NLP task. We describe the specific word embeddings that we used in our experiments.

### 2.1 Word Embeddings

Word Embeddings have proven to be efficient representation of word in several NLP tasks. Word2vec, a type of word embedding, takes as its input a large corpus of text and produces a vector space, typically of several hundred dimensions, with each unique word in the corpus being assigned a corresponding vector in the vector space. Word vectors are positioned in the vector space such that words that share common contexts in the corpus are located in close proximity to one another in the embedded vector space (Mikolov et al., 2013). In our experiments, we choose two different pre-trained word embeddings (Fasttext and M-BERT) to represent the words in Hindi language for the NER task.

Fasttext, proposed by Facebook (Bojanowski et al., 2016) provides pre-trained word embeddings of dimension 300 for Hindi (and many other languages) built on the skip-gram model, where each word is represented as a bag of character n-grams. They used Hindi Wikipedia dumps as the corpus for training the language model. They showed that the Fasttext model out performed other base line models that did not take into account sub-word information, as well as methods relying on morphological analysis. Later on, they released a pre-trained word embeddings for 157 languages including Hindi (Grave et al., 2018).

Multilingual BERT (M-BERT) was proposed by Google (Pires et al., 2019) which has an em-

bedding dimension of length 768 dimension, trained on a sub-word level using the Wikipedia corpus on multiple languages. Their model architecture consisted of a bidirectional transformer trained for the task of language modeling. They detail a novel technique they used which was called Masked LM (MLM) that enhanced the performance of their word embedding model for representation of the words as well as its use in other NLP tasks.

## 3 Dataset

We perform the task of labeling the named entities on the dataset, available at (IITH, 2008), released during ICJNLP 2008 as part of the workshop on NER for South and Sout East Asian Languages. It consists of 19822 annotated sentences, 34193 unique tokens, 490368 total tokens and 12 categories of entities and one negative entity class `other`. The 12 categories are (a) `person`, (b) `organization`, (c) `location`, (d) `abbreviation`, (e) `brand`, (f) `title:person`, (g) `title:object`, (h) `time`, (i) `number`, (j) `measure`, (k) `designation`, and (l) `terms`. चहातरो कई उचचहअ सअमअजइकअ वअ अरतहरइकअ सतअरअ का रअसायअनअ (`terms`) सअस-तरअ कए तईनो (`number`) पअसहो कए कोइ सारतहरअकअ सअमबअनदहअ नअहई हअइ is a sample sentence in the dataset. Notice that baring two words, all the others words have no tags and subsequently been considered as `other`.

We faced challenges using the above dataset, (a) 83.55% of the words in the dataset are not tagged, (b) it is not very clear if these words have not been tagged or if these are words that belong to the class `other` tag, (c) there were several sentences which had English words in the Hindi sentences, (d) inconsistencies in the tagging of various named entities namely confusion with `measure` and `number`; `time` and `number`; `designation` and `person` mostly due to the labeling being done by different people who used their own judgment to label a given word, and (e) there were several sentences which had words in parenthesis. In all there were as many as 6000 sentences, in the training set that did not have a single word in the sentence that had an entity tag. In all our experiments, we used 70% of the data to train, 15% to validate and the experimental results mentioned are on the 15% test data (the data that was

"जअहा तअकअ वासअ वइबहअगअ कए कारयअलअयअ अभइलएकहअ का सअम-बअनधअ हअइ हइमाचअलअ परअदएसहअ सअबहा कए पअतरअ तअतहअ सअमइतइ का परअतअ"

Figure 1: A sample Hindi text sentence in our test database.

not seen during the training).

### 3.1 Data Preparation

We have prepared two sets of data for our work. The first set consists of the raw data (as available from (IITH, 2008)), while the second set is produced by processing the raw data. Processing includes (a) removing all the parenthesis and the words within the parenthesis, and (b) removing all punctuation marks in the sentence. This processing is done keeping in mind the fact that the words inside parenthesis are the words that are usually not spoken and have been introduced in the text to reinforce or disambiguate a word or a concept. Also the use of punctuation marks in a sentence assumes that text is written by a person who has had a formal education in that language. The second set of data can be assumed to be the output of a speech to text engine in response to spoken sentences. Note that the second set could be looked upon as noisy text!

We experiment with both the raw data and the processed data and analyze the proposed NER models. We have also performed analysis of the data by reducing the data points. Previously as mentioned, we encounter with 60% of the data sentences that only contain words which are tagged as negative entities. So with that amount of data reduced, we have created a new dataset and also performed experiments on them. This way we can measure the robustness of the models.

## 4 Proposed approach for NER

The motivation for the proposed system is based on the following observation. A sample sentence taken from the test dataset is show in Figure 1. The following three words हइ-माचअलअ परअदएसहअ सअबहा together tagged as `organization` in our database. But the base BiLSTM model (see Figure 2) tags these as हइमाचअलअ/`location`, परअदएसहअ/`other`, सअबहा/`other`. This observation motivates us to design a modified LTSM (DAE LSTM and Condi-tional LSTM) assuming that the labels returned by the base BiLSTM could be erroneous.

$$\{w_1, w_2, \cdots w_n\} \to \boxed{\text{BiLSTM}} \to \{l_1, l_2, \cdots, l_n\}$$

Figure 2: Base Model (BiLSTM) gives a NER label to each of the $n$ words in a sentence.

We propose two new hierchical models, namely (a) Denoising Autoencoder LSTM and (b) Conditioning LSTM for improving the output of the base BiLSTM model for NER task (see Figure 2). A BiLSTM model takes in a sentence consisting of $n$ words $w_1, w_2, \cdots w_n$ and returns $n$ labels, namely, $l_1, l_2, \cdots, l_n$. Note that $l_i$ is the NER label which can take any of the 12 classes as mentioned earlier. In this paper, we consider this as the base model.

### 4.1 Denoising Autoencoder LSTM

We adopt the idea from de-noising the noisy input (Vincent et al., 2008). We consider the NER output labels obtained from the BiLSTM models, namely $l_1, l_2, \cdots, l_n$ obtained in Figure 2 are noisy. The noisy output obtained from the BiLSTM model is then provided as the input to a denoising auto-encoding (DAE) model which given a word embedding ($w_i$) and noisy tag ($l_i$) reconstructs back the word ($w_i$) and its correct ($l_i^*$) tag as shown in Figure 3. This denoising auto-encoder consists of LSTM units in the hidden layers that help in preserving the information needed to reconstruct back the data. The architecture used for the DAE consists of 2 sub-networks. The encoder and the decoder module. The encoder consists of 1 LSTM layer followed by a dense layer which compresses the input. The decoder module takes in the compressed input through a dense layer and uses a LSTM time sequence to reconstruct back the input sequence of words.

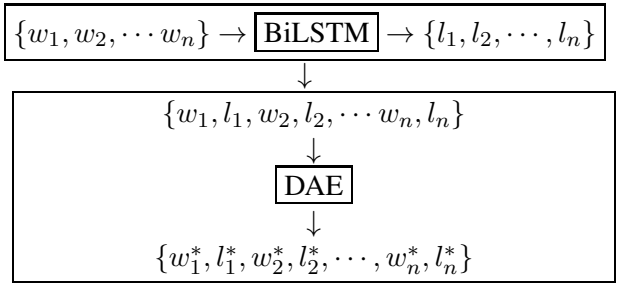

Figure 3: Denoising LSTM. Where $l_i^*$ are the true NER labels and $w_i$ are the word embeddings.

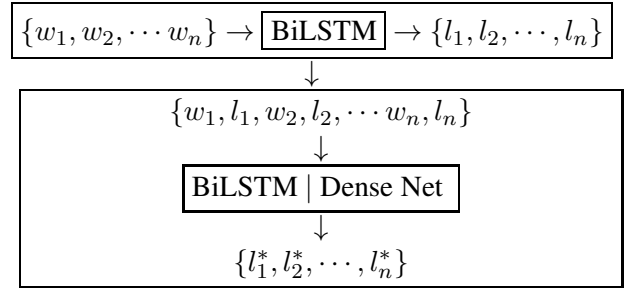

Figure 4: Conditional LSTM. Where $l_i^*$ are the true NER labels during training.

### 4.2 Conditioning LSTM

We condition the BiLSTM output (Figure 2) on the given word embedding, i.e. a multi layer perceptron or another BiLSTM network has to perform classification using the BiLSTM noisy output and the word embedding both given as an input. The idea is to use the noisy NER label information provided by the BiLSTM which can add some additional knowledge to the model along with the initial word embeddings and then perform a classification task. The LSTM conditioning module consists of 2 BiLSTM layers followed by a Dense layer which outputs the probabilities per tag. The Dense conditioning module consists of 3 Dense Layers which compress the input layer by layer to obtain a tag probability.

## 5 Experiments

We represent the words in the sentences using (a) the fasttext pre-trained Hindi word embeddings (Bojanowski et al., 2016) of size 300 and (b) BERT (Pires et al., 2019) pre-trained word embedding model of dimension 768. We have experimented with both the embeddings. Note that, in case of fasttext word embeddings, the out of vocabulary words, if any, are represented by a zero vector of size 300 while in case of BERT word embeddings, we use the word level representation for the word in the vocabulary as is and for out of vocabulary words, the BERT model finds the sub-tokens to represent the word. We take the mean of all the sub words to represent the entire word. In all our experiments we have the train-validation-test data is in the ratio 70:15:15. The number of sentences used for training was 10375 (10k data) for the first raw database and 4284 (4k data) for the processed dataset. The other hyper-parameters used for training were (a) Adam optimizer, (b) learning rate was 0.003, (c) dropout was 0.5, (d) maximum number of words in a sentence is 30 (note that $n$ in Figure 2 was 30; if a sentence had say 23 words, then $w_{24} = 0, \cdots, w_{30} = 0$ were assumed to have the embedding of 0 vector), (e) number of LSTM layers was 2. The loss function used was Categorical Cross Entropy.

| Models   | F1   | Prec | Recall |
|----------|------|------|--------|
| Fasttext | **0.71** | **0.77** | **0.66** |
| BERT     | 0.63 | 0.73 | 0.58   |

Table 1: F1 scores of BiLSTM on the test data for BERT and Fasttext embeddings.

| Batch Size | F1 |
|------------|------|
| 4  | 0.69 |
| 8  | **0.71** |
| 16 | 0.67 |
| 64 | 0.64 |

Table 2: F1 scores of various batch sizes.

| Train Data | Raw Test | Processed Test |
|------------|----------|----------------|
| 10k | **0.71** | 0.67 |
| 4k  | 0.66 | 0.62 |

Table 3: F1 scores for BiLSTM (base LSTM) Models for different training data sizes.

| Models | F1 | Prec | Recall |
|--------|------|------|--------|
| Raw dataset | | | |
| Base LSTM | 0.71 | **0.77** | 0.66 |
| DAE LSTM  | **0.72** | 0.75 | **0.69** |
| Cond LSTM (BiLSTM)   | 0.70 | **0.77** | 0.65 |
| Cond LSTM (Dense Net)| 0.71 | 0.74 | 0.69 |
| Processed dataset | | | |
| Base LSTM | 0.67 | 0.75 | 0.63 |
| DAE LSTM  | **0.72** | **0.81** | 0.65 |
| Cond LSTM (BiLSTM)   | 0.70 | 0.79 | **0.66** |
| Cond LSTM (Dense Net)| 0.71 | **0.81** | 0.64 |

Table 4: F1 score, precision and recall for various models on raw and processed dataset.

### 5.1 Results and Analysis

Initial set of experiments (see Table 1) were conducted using BiLSTM to identify the type of word embeddings and the batch size that we could use. We compared the use of Fasttext and BERT word embeddings. While BERT models (Pires et al., 2019) are meant to be used at a sentence level, however, we chose to use the embeddings at the word level. As can be seen in Table 1, the performance of the BiLSTM using Fasttext is better compared to the same model when we use BERT embeddings. We also experimented with different batch sizes while training the networks. The best results are obtained when we train the base BiLSTM network on the batch size of 8 (see Table 2). In the rest of our experiments we use Fasttext embeddings to represent the words and a batch size of 8 during training.

We have performed analysis using the hyperparameters mentioned earlier on both the 10k raw as well as 4k processed datasets. Table 3 shows the performance of the base BiLSTM model in terms of F1 score for the raw test and processed test datasets. As can be seen the performance of the BiLSTM (we will use this as the ground truth in this paper) is better when we use the raw train and test dataset. The performance is poor for processed 4k train dataset compared to the 10k raw train dataset. Even in matched train-test conditions the F1 score is 0.62 for the processed dataset. Once can attribute the degradation in performance of the BiLSTM model because of the reduced number of training samples in the processed 4k train dataset. This is along the observation that the performance of any machine learning model gets better with more training data.

The next set of experiments show the comparison of the base model (Base LSTM) with all the models that we have proposed, namely (a) Denoising LSTM (DAE LSTM), (b) Conditional LSTM (BiLSTM) and (c) Conditional LSTM (Dense Net). In all the experiments we use the 10k raw training data to build the model and use the fasttext pre-trained word embeddings, the F1 scores are on the processed test set. We can observe (see Table 4) that the proposed hierarchical model (DAE LSTM) improves on the base BiLSTM model, albeit slightly for raw 10k train dataset. Similar improvements can be seen even for the 4k processed training data, as seen in Table 4. While the base LSTM performance detoriates for the 4k processed train dataset compared to 10k raw train dataset. While we see an improvement in the F1 scores of the in-vocabulary words, we see a degradation in the performance in case of out-of-vocabulary words (see Table 5). This is expected because the word embeddings for out of vocabu-lary words is a zero vector. Subsequently, we are either conditioning the LSTM model output tags on all zero word embeddings or we are using the DAE on the word embedding (zero vector) and reconstructing back the zero vector. This does not add any additional information to the network, infact it confuses and deviates it from the target tag. In this scenario, the use of BERT word embedding or any other embeddings might prove to be useful if they can handle out of vocabulary words (Garneau et al., 2019).

| Models | F1 | Prec | Recall |
|---|---|---|---|
| Raw dataset; in-vocab words | | | |
| Base LSTM | 0.73 | **0.78** | 0.71 |
| DAE LSTM | **0.75** | 0.77 | 0.74 |
| Cond LSTM (BiLSTM) | 0.74 | **0.78** | 0.71 |
| Cond LSTM (Dense Net) | 0.72 | 0.77 | **0.70** |
| Raw dataset; OOV words | | | |
| Base LSTM | **0.31** | **0.55** | 0.24 |
| DAE LSTM | **0.31** | 0.40 | **0.27** |
| Cond LSTM (BiLSTM) | 0.25 | 0.29 | 0.26 |
| Cond LSTM (Dense Net) | 0.25 | 0.29 | 0.25 |

Table 5: F1 score, precision and recall for various models on raw data set for in vocab and out-of-vocab words.

Note that DAE LSTM outputs the word embeddings ($w_i^*$) as well (see Figure 3) as the NER label ($l_i^*$). We observed that the output of DAE LSTM resulted in a $w_i^*$ which did not exactly associate with any known word because there was not constraint placed on DAE-LSTM to output $w_i$ exactly. We hypothesize that placing such a constraint might improve the performance of DAE-LSTM. This aspect needs further investigation.

## 6 Conclusions

Named Entity Recognition is an important NLP task which has not been explored sufficiently for Hindi language. In this paper, we reused the BiLSTM architecture used for English NER by reducing the size of network to accommodate the lack of annotated data (low-resource). We proposed two extensions to the BiLSTM in the form of DAE LSTM and Conditional LSTM motivated by the fact that the NER labels output by the LSTM could be erroneous. We developed several insights into the Hindi NER problem by experimenting with different word embeddings.

Note that we wanted to avoid out of vocabulary words being represented by a zero vector (as in Fasttext), so we used BERT embeddings. However we used the BERT embeddings at the word level instead of sentence level because we we assumed that the use of BiLSTM model would capture the sentence level context. Also BERT embeddings were of dimension 768, our model (256 cell LSTM 2-layered network) couldn't accommodate embeddings of this size. Observe that if we increased the network parameters, we would require more data to train. We expect that performance of the NER system (using BERT word embeddings) will improve with either increasing the number of the layers in the network or increasing the number of cells per layer.

## 7 Acknowledgments

The work was carried out by the first authors during her internship with TCS R&I. The authors would like to thank SriHarsha Dumpala for discussion and ideas which has enhanced the experimental depth.